\documentclass[11pt,a4paper]{article}
\usepackage[hyperref]{emnlp-ijcnlp-2019}
\usepackage{latexsym}
\usepackage{graphicx}
\usepackage{multirow}

\usepackage{times}

\usepackage{url}

\aclfinalcopy 

\title{Sentence Embeddings as an intermediate target in end-to-end summarisation}

\author{Maciej Zembrzuski \\
  trivago N.V. / Düsseldorf, Germany \\
  {\tt maciej.zembrzuski@trivago.com} \\\And
  Saad Mahamood \\
  trivago N.V. / Düsseldorf, Germany \\
  {\tt saad.mahamood@trivago.com} \\}

\date{}

\begin{document}
\maketitle

\begin{abstract}
Current neural network-based methods to the problem of document summarisation struggle when applied to datasets containing large inputs. In this paper we propose a new approach to the challenge of content-selection when dealing with end-to-end summarisation of user reviews of accommodations. We show that by combining an extractive approach with externally pre-trained sentence level embeddings in an addition to an abstractive summarisation model we can outperform existing methods when this is applied to the task of summarising a large input dataset. We also prove that predicting sentence level embedding of a summary increases the quality of an end-to-end system for loosely aligned source to target corpora, than compared to commonly predicting probability distributions of sentence selection.
\end{abstract}

\section{Introduction}

Document summarisation is the task in which texts are ingested and a shorter textual summary is produced. This task continues to receive considerable attention within the Natural Language Processing community for various information access purposes and comprises of two main paradigm forms: \textit{abstractive} and \textit{extractive}. Extractive summarisation focuses on the task of choosing salient or relevant sentences in a given corpora. Whilst abstractive summarisation involves re-writing a given corpora into a more concise form that summarises the input. 

Most recent techniques for summarisation either \textit{abstractive} or \textit{extractive} have come to rely on neural network techniques with a particular focus on attempting to improve the quality of the generated summaries from long documents. This has included attempts to improve how and what text is summarised. For example, by trying to optimise extractive summarisation for the aspect of coherence \cite{Wu:2018}, for evaluation metrics \cite{Paulus:2017,Kryciski:2018, Chen:2018}, the importance of a given sentence \cite{Zhou:2018}, or by the application of a pre-trained Transformer \cite{Vaswani:2017} model as show by BERTSUM \cite{Liu:2019}. For abstractive summarisation improvements in content-selection have come about by decoupling the task of selecting of salient content and then performing abstractive summarisation as a separate discrete step \cite{Gehrmann:2018}.

However, most past work on summarisation has tended to focus on summarising newswire corpora. Work that has utilised user review corpora has either focused on generating review titles \cite{Yang:2016} or for describing the benefits or disadvantages of products \cite{Kunneman:2018}.

In this paper we present the \textsc{USEsum} model for generating short unique selling point summaries from a corpora of user hotel reviews. We use a two-stage system that combines both extractive and abstractive summarisation techniques to enable salient content selection and to allow the generation of a summary over a large input corpora. In particular, this paper makes the following key contributions:
\begin{itemize}
\itemsep0em
\item We propose a new solution to the task of summarising user reviews.
\item Demonstrate the value of representing semantic information by using pre-trained sentence embeddings through the use of the Universal Sentence Encoder \cite{Cer:2018}. 
\item Show that the use of angle measurement between sentence embeddings is a good metric for comparing the semantics of generated summary text.
\end{itemize}

\section{Related Work}
Past approaches to the problem of summarising multiple-documents have tend to be extractive in nature with the most important sentences extracted and then optionally compressed to form a summary. Dorr et al. \shortcite{Dorr:2003}, for example when generating newspaper headlines used a non-neural approach. By extracting nouns and verb phrases from the first sentence and of a news article and then using an iterative algorithm to compress the sentence to the length of a headline. Neural approaches such as the one used by Durrett et al. \shortcite{Durrett:2016} have take the same idea, but also learn a model to select sentences and compress them.

More recent approaches in document summarisation have focused on the challenge generating high quality summaries using artificial neural networks. One method to this problem has been the application of deep reinforcement learning to take into account the aspect of coherence in the generated generated summaries \cite{Wu:2018} or to directly optimise for evaluation metrics \cite{Paulus:2017, Kryciski:2018, Chen:2018} to improve the quality of the generated summarised output.

Alternative approaches for improving the quality of extractive summarisation have included the use of a pre-trained \textsc{BERT} \cite{Devlin:2018} model such as the \textsc{BERTSUM} system \cite{Liu:2019}. This approach predicts a score for each sentence directly in addition to using the Transformer model on-top of sentence embeddings. This has resulted in improved \textsc{ROUGE} scores when tested on \textit{CNN/Daily Mail} corpora in comparison to other state-of-the-art extractive systems. Systems such as \textsc{NEUSUM} \cite{Zhou:2018} demonstrate a successful approach in the use of model sentence importance scores and thus improving the quality of its extractive summarisation. Additionally, both the  \textsc{REFRESH} \cite{Narayan:2018} and \textsc{DCA} \cite{Celikyilmaz:2018} systems also show that the summarisation task can improve with the application of reinforcement learning. 


Abstractive summarisation attempts to to produce a summary, which may contain aspects that were not part of the original input. The use of a neural language model in combination with a beam search decoder can result in the generation of more accurate summaries \cite{Rush:2015}. Nevertheless, one of the problems with end-to-end neural network-based methods for abstractive summarisation tend to perform poorly at content selection \cite{Gehrmann:2018}. This is because such models can include content that is neither relevant or salient in the generated summary. One technique to fix this deficiency has been the development of a two-stage generation approach. Firstly by performing content-selection through the use of a bottom-up selector that selects salient phrases in the source document and secondly by performing the step of generating of abstractive summaries using a standard neural model with the given selection mask. This two step-approach has show to outperform other alternative approaches that have attempted to perform content-selection as part of an end-to-end model \cite{Gehrmann:2018}. 

An alternative to the two-stage abstractive summarisation technique proposed by Gehrmann et al. \shortcite{Gehrmann:2018}, is the use of a discourse parser to obtain a discourse tree of user products reviews. Selection of content is driven by the use of the PageRank algorithm to select a sub-graph of the most important aspects, which in combination with a template based Natural Language Generation is used to generate an abstractive summary \cite{Gerani:2014}.

An additional challenge is the ability to generate high quality abstractive summaries when faced with large documents or summaries without including repetitive or incoherent phrases. Work by Paulus et al. \shortcite{Paulus:2017} has shown that with the use of a intra-attention model that pays attention to both previously used input tokens in the encoder and the words already generated by the decoder it is possible to generate higher quality and more readable abstractive summaries \shortcite{Paulus:2017}.

Most past work related to utilising user reviews for summarisation has focused on predicting review titles \cite{Yang:2016}. The work of Kunneman et al. \shortcite{Kunneman:2018}, for example, focused on extracting the positives and negatives of products from user reviews. This approach required the reviews to be formatted in a specific structure, which is not present in the analysed dataset.  Overall, the reviews summarisation task differs significantly from news summarisation due to the phenomena of opinion shifting  \cite{Pecar:2018} and the lack of document structure.

\section{Unique Selling Point dataset}

The USEG dataset\footnote{USEG — https://github.com/useg-data/useg-data} represents the problem of describing the unique characteristics of a hotel based on user reviews. Each target hotel description consists of one sentence and the related reviews contain up to 800 sentences per hotel. 

This problem can be seen as a multi-document summarisation challenge as each user reviews is independent of each other. On the other hand, reviews are short and written in various styles. This limits the possibility of benefiting from document structure. In this approach the reviews were merged into a single document per hotel. Therefore, the summarisation tasks depends more on extracting the most interesting information, irrespective of where the information is located in the input text. We see this task as a special kind of single document summarisation challange where there is no document structure. 

The task is also unique in the way that the summaries are not necessarily meant to cover the most commonly represented features or amenities of a hotel. For example, a hotel can be well located, but what makes it unique is the fact that it contains a rooftop swimming pool. The “compression rate” is relatively high compared to commonly used datasets  (e.g. \textit{CNN/Daily Mail}, \textit{DUC 03/04}, \textit{Gigaword}, etc.). This dataset requires the analyse of up hundreds of sentences to generate a single sentence. The style of descriptions also differ significantly from the reviews’ style, therefore the word overlap between the descriptions and the user reviews is smaller than in commonly used datasets. In the USEG dataset only a third of the summaries overlap sufficiently with source reviews that the \textsc{NEUSUM} \shortcite{Zhou:2018} approach could use them for training, based on the \textsc{ROUGE} score \cite{Lin:2004}. However, when similarities were calculated using pre-trained sentence embeddings \cite{Cer:2018}, two thirds of the descriptions contained information covered by reviews that were formulated using similar but different expressions. The remaining third of summaries don’t contain information which is reflected in aligned user reviews. Nevertheless, this is still valid for training the extractive approach in proposed architecture.

\section{Proposed Approach}

Due to the nature of the \textsc{USEG} dataset, where the target sentences differ significantly in style and length from input sentences, the task requires an abstractive approach to generate the USP summaries. On the other hand, due to large amount of input sentences, it is easier to retrieve the requisite information about the unique characteristics of a given accommodation in an extractive manner from the user reviews. Therefore, the \textsc{USEsum} system consists of both approaches.  An extractive model firstly selects the top three sentences from user reviews while a secondary abstractive approach predicts the final summary description. The inspiration for selecting three sentences for further processing is the fact that selection of the three initial sentences (LEAD 3 \shortcite{Nallapati:2016}) is sufficient to constitute a strong baseline in \textit{CNN/Daily Mail} summarisation.

\begin{figure}[h]
  \includegraphics[width=\linewidth]{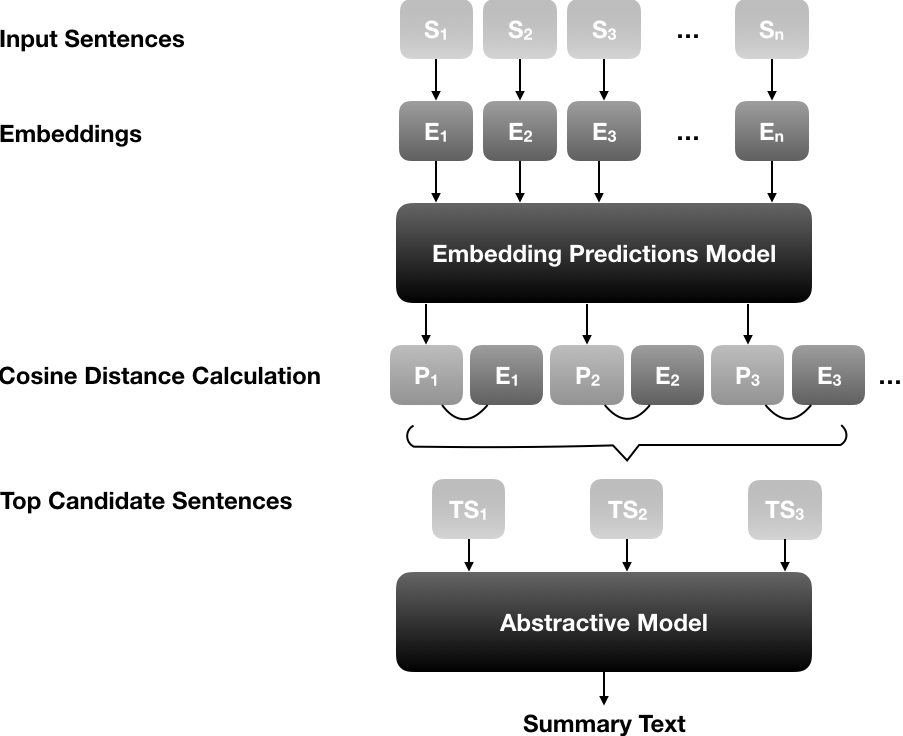}
  \caption{\textsc{USEsum} system diagram for generating USP abstractive summaries}
  \label{fig:sysdiagram}
\end{figure}

To further limit processing on the large amount of input sentences by the extractive model, they are pre-processed in such a way that each input sentence is represented by a semantically meaningful vector, furtherly referred to as a sentence embedding. The sentence to vector calculation is done using the Universal Sentence Encoder \cite{Cer:2018}. The extractive model predicts sentence embedding vectors in the same space as the input vectors. This allows for the selection of the most relevant sentences from the given input, which are concentrated around a common concept. After predicting the sentence embedding, it is compared with the embeddings of the input sentences and the three most similar input sentences are selected. This proposed system was implemented using a custom adaptation of \textsc{OpenNMT} \cite{Klein:2017}. Figure \ref{fig:sysdiagram} illustrates an abstract overview of how the \textsc{USEsum} system performs end-to-end summarisation.

The source code and models used for the USP summary generation task are available online\footnote{USP summary generation implementation code and models — https://github.com/USE-sum/usesum}.

\subsection{Problem definition}

Extractive document summarisation aims to select the most salient sentences from a given document. Each document $D_{j}$ in set of documents to summarise, contains $L$ sentences $D_{j}=(S_{1}, S_{2},..S_{i}.., S_{L})$. The extractive summarisation system tries to identify the subset of $D_{j}$ which in the \textsc{USEsum} setting will be further processed by the abstractive summarisation system. 

During the training phase of the extractive system, the sentence embedding of the reference summary $y$ and the embedding $x_{i}$ of the $i-th$ processed sentence are given. Both $y$ and $x_{i}$ are semantically meaningful sentence embeddings, calculated by Universal Sentence Encoder \cite{Cer:2018}. The goal of training is to learn a scoring function which will predict the influence of the currently processed input embedding $x_{i}$ on the final embedding prediction $\hat{y}_{L}$ of the document $D_{j}$. 

The proposed approach follows Maximal Marginal Relevance (MMR) method proposed by Carbonell and Goldstein \shortcite{Carbonell:1998}. The MMR method tries to maximise the relative gain given previously extracted information. In the proposed approach the information gain at $i-th$ step is measured by comparing angles between sentence embeddings $y$ and $\hat{y}_{i}$. 

The \textsc{USEsum} model is trained to maximize a scoring function $g(·)$ of the information gain represented by:
\begin{equation}
g(x_{i}|\hat{y}_{i-1}) = angle(y,\hat{y}_{i}) - angle(y,\hat{y}_{i-1})
\label{gain_func}
\end{equation}
At each processing step $i$ the summarisation system estimates the influence of the processed input sentence embedding $x_{i}$ on the predicted document embedding $\hat{y}_{L}$.

\subsection{Extractive approach}

Considering the lack of structure of the input documents in the dataset, we have employed the use of the Universal Sentence Encoder \cite{Cer:2018} to transform the documents represented as a sequence of sentences into a sequence of sentence embeddings. These sentence embeddings are further encoded by the bi-directional \textsc{LSTM} \cite{Hochreiter:1997} resulting in encoder memory states at $i-th$ step represented by $s_{i}$. 
Subsequent processing by the decoder is defined by the following equations:
\begin{equation}
fg = \sigma(W_{fg}([s_{i}, \alpha , \beta ])
\label{fg}
\end{equation}
\begin{equation}
ig = \sigma(W_{ig}([s_{i},\alpha , \beta ])
\label{ig}
\end{equation}
\begin{equation}
og =  \sigma(W_{oxg}(x_{i}) + W_{ogy}(\hat{y}_{i-1})) 
\end{equation}
\begin{equation}
ct = fg * \hat{y}_{i-1} + ig * tanh(W_{ct}(x_{i}))
\end{equation}
\begin{equation}
ht = og * \sigma(ct)
\end{equation}
\begin{equation}
score = \sigma(W_{sc}(ht))
\end{equation}
\begin{equation}
\hat{y}_{i} = \hat{y}_{i-1} - score * (\hat{y}_{i-1} - x_{i})
\end{equation}

Where \textit{W} are learnable weights, $\sigma$ is a sigmoid transformation, $\hat{y}_{i}$ is the prediction at ${i}$-th processing step, and $x_{i}$ is the sentence embedding of the $i-th$ input sentence. $\alpha$ is the cosine similarity between $\hat(y)_{i-1}$ and $x_{i}$ . $\beta$ is the cosine similarity between $\hat(y)_{i-1}$ and vector $\omega_{j}$ representing all sentences in a document. $\omega_{j}$ is calculated by the following equation: 
\begin{equation}
\omega = \sum_{i=0}^{N_{j}}x_{i}
\label{omega}
\end{equation}
For processing of the first sentence, $\hat{y}_{i-1}$  is initialised as:
\begin{equation}
\hat{y}_{-1}  = tanh(W(tanh(\omega _{j}))))
\label{prev_pred}
\end{equation}
 Scoring the influence of the difference between $x_{i}$ and $\hat{y}_{i-1}$ aims at mitigating the phenomena of information redundancy and dealing with the fact that the most popular information doesn't have to be related the target. The decision to focus on the use of a RNN instead of using a transformer was motivated by the very long inputs, which resulted in reaching GPU memory constraints. 

\subsubsection{Objective function}
A training step is defined as the processing of a single input sentence and predicting new estimation of the target vector. The loss calculation at each training step depends on the arcus cosine value of cosine similarity between predicted embedding vector and the target embedding. The final loss at each step is the difference between the current and previous step distances from the target. Therefore, there is no loss during the first training step. The loss is negative when an improvement in comparison to previous estimation has occurred. We found that training the model with both positive and negative loses improves performance.

The model is expected to optimize the information extraction for each sentence. Predicting the final vector is a side effect of this approach. 
\begin{equation}
    loss_{i} = acos(\hat{y}_{i-1} ,target) -acos(\hat{y}_{i}, target)
\end{equation}

Where $\hat{y}_{i}$ is the sentence embedding prediction at step ${i}$. $acos$ stands for arcus cosine similarity.
This approach allows for utilisation of training examples where there is a week alignment between sources and target. The model can still learn which sentences are comparatively more informative, even if none of them matches the target perfectly.

\subsection{Abstractive summarisation} \label{abs_sum}

The predicted embedding $\hat{y}$ is further used to extract three sentences which are an input for abstractive summarisation. These sentences are selected based upon the angle similarity between each of the input sentence embeddings $x_{i}$ and the predicted summary embedding $\hat{y}_{L}$.
For abstractive summarisation we have employed the use of a Transformer \shortcite{Vaswani:2017} in a standard configuration with additional word features such as word lemma, POS tags, NER tags, and dependency type. This was inspired by the work of Nallapati et al. \shortcite{Nallapati:2016}. Word embeddings are initialised using 300 dimensions of \textsc{GloVe}\footnote{\textsc{GloVe} — https://nlp.stanford.edu/projects/glove/}  \cite{Pennington:2014} and the copy attention proposed by Gu et al. \shortcite{Gu:2016}. The loss function was altered by the usage of focal loss \cite{Lin:2017} with the default parameter \textit{$\gamma$=2}. We also updated beam search in the inference phase to recognise potential named entities in the generated summary candidates, to penalise entities which were not present in source text and promote weights of the candidate entities present in the input. As discussed more broadly in the Results section, this approach helped to solve problematic cases such as a wrong city would be mentioned in a generated summary. The penalty and promotion factors were estimated manually on validation set. The beam scores of candidate words recognised as named entities, not present in the input, are multiplied by a factor of \textit{50}, and the ones present by a factor of \textit{0.4}.
The abstractive model is trained independently from the extractive model. During training phase, the model is provided with three of the most similar sentences measured by the angle similarity between these sentences' embeddings and target summary embedding.

\section{Evaluation} \label{eval}

To evaluate the proposed solution for generating USP summaries we have chosen the USEG dataset. This dataset is to the best of our knowledge the only publicly available dataset for summarising user reviews at a product level. This dataset differs considerably from usually used datasets for document summarisation, such as \textit{CNN/Daily Mail}, \textit{NYTimes}, \textit{DUC-03}, and \textit{DUC-04} \cite{Hermann:2015,Sandhaus:2008,Over:2003,Over:2004}. The main features differentiating the USEG dataset from the commonly recognised datasets are the following aspects:

\begin{itemize}
	\item Lack of document structure.
	\item Random distribution of information.
	\item Single sentence summaries.
	\item Low word overlap between source and target texts.
\end{itemize}

We have optimised the model to tackle these inherit challenges in the USEG dataset due to consisting of user generated reviews of accommodations. Adapting this approach for other datasets would required significant changes in architecture, which was considered out of scope of for this project due to time constraints. The changes would consist of adaptations to allow the model to benefit from the presence of a document structure and to be able to generate coherent multi-sentence summaries.

In addition to choice of dataset, we have have also chosen commonly recognised metrics for evaluating NLP based summarisation implementations such as \textsc{ROUGE-L}, \textsc{BLEU}, and \textsc{METOR}. We also used cosine similarity between sentence embeddings of generated summaries and targets as this would measure the semantic similarity between source and target irrespective of the words chosen. This is because in the proposed solution we have utilised the sentence embeddings calculated by the Universal Sentence Encoder \cite{Cer:2018} and these embeddings are semantically meaningful to make the cosine similarity a good measure for comparing similarity.    

To evaluate USEsum model, we compare it with pre-existing competitive summarisation systems that have utilised the \textit{CNN/Daily Mail} dataset. These systems were chosen on the basis of their performance on \textit{CNN/Daily Mail} dataset, whether we were able to find their implementation code, and whether we could adapt the systems within the given time constraints to the \textsc{USEG} dataset. The chosen extractive systems include \textsc{REFRESH} \cite{Narayan:2018},  \textsc{BERTSUM} \cite{Liu:2019}, and \textsc{NEUSUM} \cite{Zhou:2018}. For these extractive models, we made minor adaptations to enable processing of USEG dataset. Maximal input sentence size for \textsc{NEUSUM} was \textit{800} sentences. For \textsc{BERTSUM} we retained a \textit{512} input word limit. This limited its performance due to the fact it could only process a fraction of the input data. However, changing this parameter would require deeper changes in it's implementation code. We added a baseline Universal Sentence Encoder \cite{Cer:2018} approach (further: BASELINE) where all the input sentence embedding vectors of a document were summed to $\omega$ vector as defined in equation \ref{omega}. Furthermore, the sentence with the embedding that is most similar to the $\omega$ is selected.

For performing end-to-end comparisons the outputs of all the extractive models were processed by a common abstractive model. 

For a comparison with an alternative end-to-end abstractive summarisation model we also trained a RNN model using \textsc{OpenNMT}. This was done using the pre-existing BOTTOM-UP abstractive summarisation approach \cite{Gehrmann:2018} from the documentation provided on the \textsc{OpenNMT} website\footnote{OpenNMT Summarisation Documentation — http://opennmt.net/OpenNMT-py/Summarization.html}. The embeddings were initialized with \textit{300} dimensional \textsc{GloVe} pre-trained embeddings. The encoder hidden size was \textit{512} and decoder \textit{1024}. To make the comparison more fair, the input words are augmented with the same features as described in section $\ref{abs_sum}$.

\section{Results}
To evaluate the effectiveness of \textsc{USEsum}, we performed several independent experiments, aiming to evaluate the following aspects:
\begin{itemize}
\item Evaluating the quality of the extractive summarisation.
\item Evaluating the quality of the summary generated end-to-end.
\item Estimating the effect of beam search input word promotions as described in section \ref{abs_sum}.
\end{itemize}

The inferred results for all of the experiments are available online\footnote{Experimental Results — https://github.com/USE-sum/usesum/tree/master/results}.

\subsection{Extractive summarisation experiment} \label{extsum}
The first experiment was to compare the quality of prediction of the best sentence selected by extractive models. Table 1 presents the results of the first experiment. The models selected for comparison are described in section $\ref{eval}$. This experiment shows interesting phenomena; \textsc{NEUSUM} outperforms other approaches in all metrics except for cosine similarity between embeddings. The latter is interesting as it is the only metric which measures the semantic similarity of sentences, irrespective of the words used. As the final experiment in section $\ref{end2end}$ shows the end-to-end \textsc{USEsum} model outperforms other end-to-end approaches, which rely on extractive models for all metrics. This suggests the high importance of cosine metrics in this case. 

\begin{table*}[h!]
  \begin{center}
  \begin{tabular}{|l|c|c|c|c|l}
    \hline
    &BLEU&ROUGE-L&METEOR&Cosine Similarity \\
    \hline
    BASELINE&0.0027&0.0632&0.0358&0.3874 \\
    \textsc{BERTSUM} 1$^{st}$ sentence&0.0011&0.0503&0.0275&0.3502 \\
    \textsc{NEUSUM} 1$^{st}$ sentence&\textbf{0.0055}&\textbf{0.0846}&\textbf{0.0539}&0.4149 \\
    \textsc{REFRESH} 1$^{st}$ sentence&0.0004&0.0459&0.0227&0.2965 \\
    \hline
    \textsc{USEsum} 1$^{st}$ sentence&0.003&0.0761&0.0474&\textbf{0.4493} \\
    \hline
  \end{tabular}
  \end{center}
  \caption{Extractive summarisation results}
  \label{tab:extractivesummresults}
\end{table*}

\subsection{End to end summarisation experiment} \label{end2end}
For the end-to-end approach, aiming at generating the desired summaries, we compare different extractive based models by combining them with the same abstractive model to obtain the final summary. To compare, we also include the \textsc{BOTTOM-UP} \cite{Gehrmann:2018} abstractive end-to-end model. All the models generated ten summary candidates. To select the best candidate, we used a simple heuristics of comparing each candidate sentence embedding with the embedding of the three input sentences for the abstractive model. For the \textsc{BOTTOM-UP} model the whole document was used as the input, therefore the candidates were compared with the embedding of the whole document. For additional comparison we also used a BOTTOM-UP model in which only the first sentence candidate is chosen as the final summary. This system is referred to as {BOTTOM-UP} $1^{st}$. 
The results of the end-to-end experiment are shown in table \ref{tab:endtoendresults}. 

\begin{table*}[h!]
  \begin{center}
  \begin{tabular}{|l|c|c|c|c|l}
    \hline
    &BLEU&ROUGE-L&METEOR&Cosine Similarity \\
    \hline
    \textsc{BASELINE}&0.0063&0.1030&0.0448&0.489\\ 
    \textsc{BERTSUM}&0.0040&0.1071&0.0414&0.4924 \\
    \textsc{BOTTOM-UP} &0.0188&0.1427&0.0543&\textbf{0.5132} \\
    \textsc{BOTTOM-UP} $1^{st}$ &0.0153&0.1213&0.0468&0.4937 \\
    \textsc{NEOSUM}&0.0208&0.1217&0.0535&0.4866 \\
    \textsc{REFRESH}&0.0044&0.0948&0.0379&0.4559 \\
    \hline
    \textsc{USEsum}&\textbf{0.0225}&\textbf{0.1479}&\textbf{0.0602}&0,5115 \\
    \hline
  \end{tabular}
  \end{center}
  \caption{End to end summarisation results}
  \label{tab:endtoendresults}
\end{table*}

Both \textsc{BERTSUM} and \textsc{REFRESH} omit some test cases where they were unable to decide which sentence to choose. In total, the number of omitted cases amounted to two for BERTSUM and six for REFRESH. 
The comparison between {BOTTOM-UP} and {BOTTOM-UP} $1^{st}$ shows that using the Universal Sentence Encoder for selecting the best candidate, instead of picking the top output from the system, is beneficial. This post selection of candidates also improved results for all other models in additional experiments, which is not listed here for brevity. 
The USEsum model seems to outperform other models considering all metrics with the exception of cosine similarity where the BOTTOM-UP model performs best. The human analysis of the results gives more insights into the results and cosine metrics.

\subsection{Human evaluation of the end-to-end experiment}
In addition to the evaluation with automatic metrics, a manual evaluation was also conducted to estimate the quality and semantics of outputs generated by the \textsc{BOTTOM-UP}, \textsc{NEUSUM}, and \textsc{USEsum} models in comparison to the target summary. Table \ref{tab:exampleextractiveoutputs} illustrates the example outputs from these three systems.

The predictions from each system were given to human evaluators who performed an intrinsic evaluation for a hundred outputs on each system. The evaluators marked the number of outputs that were grammatically correct and whether the output covered the semantics of the target or not. Unlike the binary ratings assigned for grammatical correctness, if the output text partially covered the target semantics, for example mentioning one amenity or facility in a given target but not others, then the output text could be awarded half a point. 

Table \ref{tab:e2eoutputshuman} illustrates the results of this evaluation between the three systems. Whilst, the \textsc{BOTTOM-UP} system performed best in terms of grammatical correctness, \textsc{USEsum} was better than other systems for semantic similarity with the intended target text. The \textsc{BOTTOM-UP} system learned to use several popular phrases and generated outputs by combining these phrases. For example, it mentioned the phrase ``pet friendliness" in 96 predictions whereas there were only 9 in the target summaries. This approach helps in keeping grammatical correctness, however it also results in predicting features which are not present in the summarised document.

\begin{table*}[ht]
\resizebox{\textwidth}{!}{\begin{tabular}{|c|l|l|l|l}
\cline{1-4}
\textbf{Example} & \multicolumn{1}{c|}{\textbf{Target}}         & \multicolumn{1}{c|}{\textbf{System}} & \multicolumn{1}{c|}{\textbf{Inference}}                &  \\ \cline{1-4}
                 &                                              & USESum                               & Great location in the city                             &  \\ \cline{3-4}
1                & Communal computer available in the lobby     & NEUSUM                               & Offers a flat lounge and free computer room for guests &  \\ \cline{3-4}
                 &                                              & BOTTOM-UP                              & Free Wi-Fi and continental breakfast                   &  \\ \cline{1-4}
                 &                                              & USEsum                               & Rooftop terrace with view over the beach               &  \\ \cline{3-4}
2                & Rooftop terrace with fantastic views         & NEUSUM                               & Fitness centre with rooftop terrace                    &  \\ \cline{3-4}
                 &                                              & BOTTOM-UP                              & Pet-friendly hotel with full kitchens                  &  \\ \cline{1-4}
                 &                                              & USESum                               & Stylish rooms with lovely wooden floors                &  \\ \cline{3-4}
3                & Funky, modern décor in the heart of Valencia & NEUSUM                               & Spacious rooms with extra wooden floors                &  \\ \cline{3-4}
                 &                                              & BOTTOM-UP                              & Modern hotel with free breakfast                       &  \\ \cline{1-4}
\end{tabular}}
  \caption{Example extractive outputs from \textsc{USEG}, \textsc{NEUSUM}, and \textsc{BOTTOM-UP} systems}
  \label{tab:exampleextractiveoutputs}
\end{table*}

\begin{table*}[ht]
\begin{center}
\begin{tabular}{|l|c|c|}
\hline
\multicolumn{1}{|c|}{System} & Grammar \% & Semantics \% \\ \hline
USEsum                       & 60                 & \textbf{10}       \\ \hline
NEUSUM                       & 56                 & 8                 \\ \hline
BOTTOM-UP                    & \textbf{91}        & 7.5               \\ \hline
\end{tabular}
\end{center}
\caption{Human evaluation of end-to-end outputs}
\label{tab:e2eoutputshuman}
\end{table*}

\subsection{Input word promotion in beam search} \label{beamsearch_promotion}
To measure the influence of promoting input named entities and nouns during beam search of the abstractive model, we repeated the end-to-end experiment with the input promotions turned off and all other parameters left unchanged.  The results of the inference without this feature are shown in table \ref{tab:beamsearchresults}. The comparison with results in table \ref{tab:endtoendresults} shows improvements in all metrics for all compared models when using input promotions. 

\begin{table*}[h!]
\begin{center}
  \begin{tabular}{|l|c|c|c|c|c|l}
 \hline
 & BLEU & ROUGE-L & METOR & Cosine Similarity \% \\ 
  \hline
{\small BASELINE, No Promotion} &0.0059&0.0761&0.0265&0.4705 \\ 
{\small \textsc{NEUSUM}, No Promotion} &0.0075&0.0838&0.0332&0.4396  \\ 
{\small USEsum, No Promotions} &0.0104&0.0875&0.0337&0.4482 \\ 
 \hline
\end{tabular}
\end{center}
  \caption{Beam search without promoting source words as output}
  \label{tab:beamsearchresults}
\end{table*}

\subsection{Assessment of cosine similarity metrics}
Cosine similarity between pre-trained sentence embedding measures the semantic relatedness between these sentences irrespective of word overlap. Therefore, this is a promising proxy for comparing quality of generated summaries. As shown in section \ref{extsum}, this cosine metric was the only one to predict the best performance of the USEsum model in the end-to-end evaluation. However, depending solely on angle similarities between pairs of embeddings may be misleading, as the same angle may be calculated for points in very different locations in the semantic embedding space. This results in the assignment of higher cosine similarity scores to the models that excel in averaging predictions. This is demonstrated by the BOTTOM-UP approach outperforming USEsum in cosine similarity as shown in table \ref{tab:endtoendresults}. This is despite BOTTOM-UP generating repetitive predictions. A similar observation can be made in the beamsearch experiment (section \ref{beamsearch_promotion}). Results in table  \ref{tab:beamsearchresults} show that cosine similarity metrics is useful for comparing similar models (USEsum and NEUSUM) but overly promotes the BASELINE approach which simply averages vectors in a document. Therefore, we can claim that cosine similarity is a valuable metric for assessing the outputs from similar systems.

\section{Conclusion}
We have proposed a novel method for summarising large unstructured documents containing various styles into short uniform styled summaries. The proposed approach outperforms competitive solutions as measured in the USEG-based evaluation.

We found that using sentence embeddings, calculated by Universal Sentence Encoder \cite{Cer:2018}, for measuring information gain and similarity is beneficial when processing texts with low levels of word overlap. The results we obtained confirm the semantic meaningfulness and high accuracy of these vectors. 

We also showed that by using simple heuristics for adjusting beam scores of word candidates, improves the end-to-end summarisation task with the USEG dataset.

\section{Future Work}
A future extension of the \textsc{USEsum} system would include adaption and evaluation for the \textit{CNN/Daily Mail} newswire corpora. Currently, the decoder of the extractive model is \textsc{LSTM} inspired. We would like to adapt the universal transformer as a decoder in the extractive model. Our motivation is the fact that the decoder based on a transformer used in \textsc{BERTSUM} outperformed the \textsc{RNN} decoder. 

Other possibilities include adapting USEsum for other multi-lingual summarisation, beyond English, by using universal language-agnostic sentence level embeddings through implementations such as Facebook's \textsc{LASER}\footnote{Facebook \textsc{LASER} — https://code.fb.com/ai-research/laser-multilingual-sentence-embeddings/}. 

\newpage
\bibliography{emnlp-ijcnlp-2019}
\bibliographystyle{acl_natbib}

\appendix


\end{document}